\pdfoutput=1

\documentclass[11pt]{article}

\usepackage[preprint]{acl}

\usepackage{times}
\usepackage{latexsym}

\usepackage[T1]{fontenc}

\usepackage[utf8]{inputenc}

\usepackage{microtype}

\usepackage{inconsolata}

\usepackage{graphicx}

\usepackage{tikz}
\usepackage{lipsum} 
\usepackage{amsmath}
\usepackage{arydshln}
\usepackage{cleveref}
\usepackage{listings}
\crefname{section}{§}{§}
\crefname{appendix}{Appendix}{§}
\usepackage{caption}
\usepackage{subcaption}
\usepackage{booktabs}
\usepackage[most]{tcolorbox}
\usepackage{amssymb}

\title{
SampleMix: A Sample-wise Pre-training Data Mixing Strategey by Coordinating Data Quality and Diversity}

\author{Xiangyu Xi$^{1}$\footnotemark[1] , Deyang Kong $^{1,2}$\footnotemark[1], Jian Yang$^{1}$\footnotemark[1], JiaWei Yang$^{1}$, Zhengyu Chen$^{1}$, Wei Wang$^{1}$, \\
\textbf{Jingang Wang$^{1}$}, \textbf{Xunliang Cai$^{1}$}, \textbf{Shikun Zhang$^{2}$}, \textbf{Wei Ye$^{2}$\footnotemark[2]} \\
  $^1$ Meituan Group, Beijing, China \\
  $^2$ National Engineering Research Center for Software Engineering, Peking University, \\Beijing, China \\
   \texttt{xixy10@foxmail.com, wye@pku.edu.cn} \\}

\begin{document}
\maketitle
\footnotetext[1]{The first three authors contributed equally.}
\footnotetext[2]{Corresponding authors.}
\begin{abstract}
Existing pretraining data mixing methods for large language models (LLMs) typically follow a domain-wise methodology, a top-down process that first determines domain weights and then performs uniform data sampling across each domain.
However, these approaches neglect significant inter-domain overlaps and commonalities, failing to control the global diversity of the constructed training dataset.
Further, uniform sampling within domains ignores fine-grained sample-specific features, potentially leading to suboptimal data distribution. 
To address these shortcomings, we propose a novel sample-wise data mixture approach based on a bottom-up paradigm. This method performs global cross-domain sampling by systematically evaluating the quality and diversity of each sample, thereby dynamically determining the optimal domain distribution.
Comprehensive experiments across multiple downstream tasks and perplexity assessments demonstrate that SampleMix surpasses existing domain-based methods. 
Meanwhile, SampleMix requires 1.4x to 2.1x fewer training steps to achieves the baselines’ performance, highlighting the substantial potential of SampleMix to optimize pre-training data.

\end{abstract}

\section{Introduction}
\label{sec:introduction}

\begin{figure}[!hbt]
    \centering
    \includegraphics[width=1.0\linewidth]{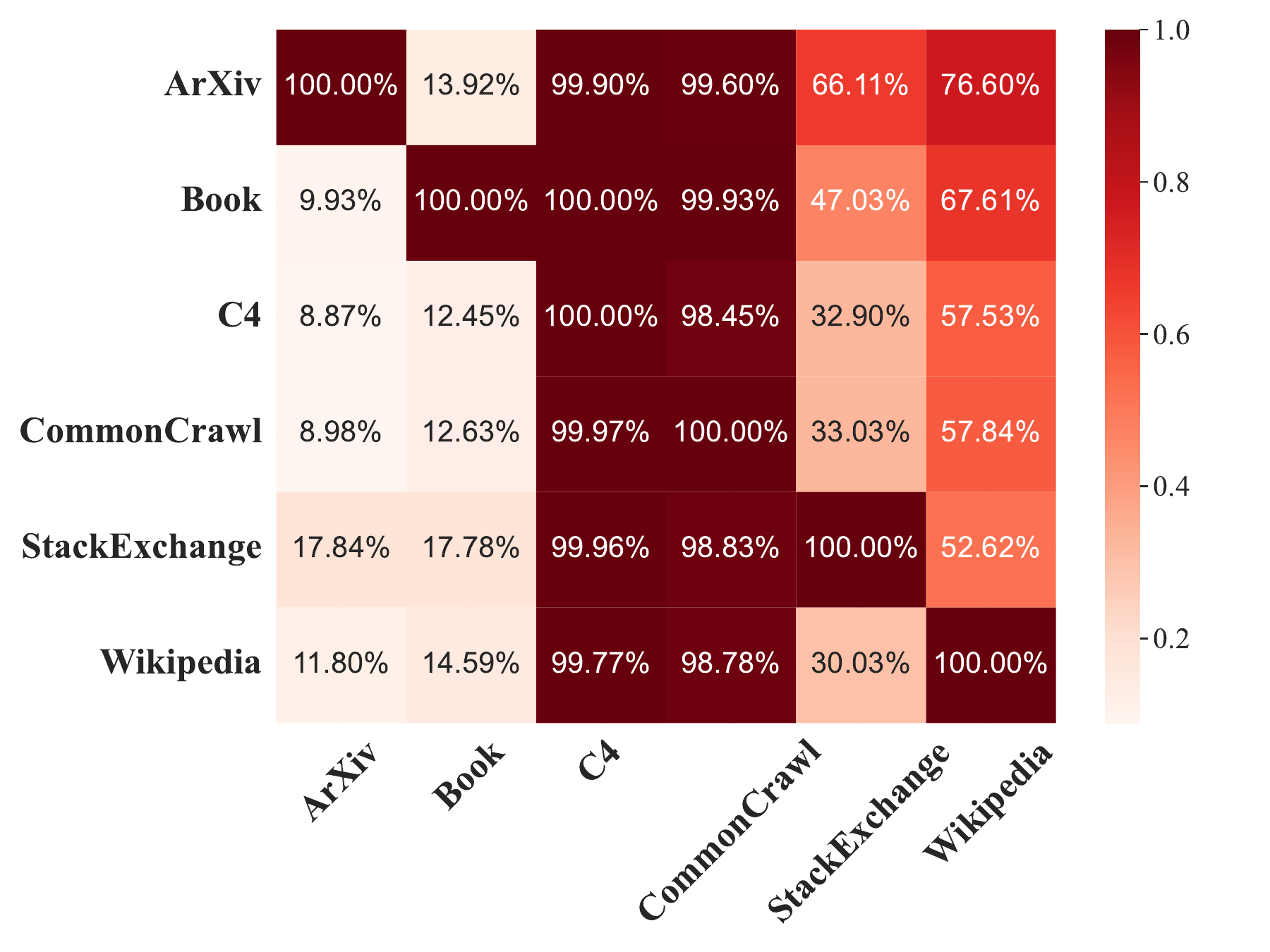}
    \caption{
    We conduct data clustering analysis using the SlimPajama dataset.
    For each domain (row), each cell shows the percentage of its clusters that also include samples from other domains (column). E.g., 76.60\% of ArXiv's clusters include Wikipedia samples (1st row, 6th column). The results reveal substantial overlap between domains.
    }
    \vspace{-4mm}
    \label{fig:cluster_overlap}
\end{figure}





The mixture proportions of pretraining data, which greatly affect the language model performance, have received increasing attention from researchers and practitioners.
In the early years, heuristic-based methods were widely employed to assign domain weights using manually devised rules, such as upsampling high-quality datasets (e.g., Wikipedia) multiple times \cite{gao2020pile,laurenccon2022bigscience}.
Afterwards, models like GLaM \cite{du2022glam} and PaLM \cite{chowdhery2023palm} established mixture weights based on the performance metrics of trained smaller models.
More recently, learning-based methods have been proposed, involving the training of small proxy models across domains to generate optimal domain weights \cite{fan2023doge,xie2024doremi}.
These existing methods follow a domain-wise methodology, a top-down process that first determines the proportion of each domain and then samples uniformly from the selected domain. Despite achieving advancements, These approaches present two key issues: 

\textbf{(1) Ignoring Inter-domain Overlaps and Commonalities.} In current pretraining datasets, ``domain'' is primarily categorized based on data sources rather than intrinsic textual or semantic properties. An implicit assumption of the domain-wise approaches is that samples are distinct and unrelated across domain boundaries. However, in practice, samples across different domains exhibit significant shared characteristics, both in terms of raw text and high-level semantics. 
To examine this assumption, we analyzed the SlimPajama dataset \cite{cerebras2023slimpajama}, a quality-filtered and deduplicated dataset, focusing on relationships between samples and clusters across its six text domains (excluding GitHub). For each domain, we computed the percentage of its clusters that also included samples from other domains, as Figure \ref{fig:cluster_overlap} shows. Our findings reveal substantial overlap between domains—nearly all clusters contain samples from both CommonCrawl and C4. Furthermore, manual inspection of the clustered samples confirms that data from different domains frequently share similar topics and characteristics. For instance, Figure \ref{fig:samples_of_same_topic} illustrates that samples from multiple domains discuss Einstein and the Theory of Relativity. By disregarding inter-domain commonalities, domain-wise mixture methods fail to control the global diversity of training data effectively. 

\textbf{(2) Suboptimal Sample Distribution within Domains.} A second limitation arises from the uniform sampling within each domain, which can lead to a suboptimal distribution of training samples \cite{xie2024doremi,fan2023doge,ye2024data}. Intuitively, samples with higher quality and greater diversity should have a higher probability of being selected \cite{xie2023data,abbas2023semdedup}. At the same time, lower-quality samples should not be entirely discarded, as they contribute to the model's generalization ability \cite{sachdeva2024train}. Determining an effective sampling strategy within each domain is nontrivial, yet current approaches lack fine-grained control over sample selection.

To address these limitations, we propose a novel sample-wise data mixture approach with a bottom-up paradigm. Instead of defining domain proportions upfront, we first perform global sampling across the dataset based on sample quality and diversity, dynamically determining domain distributions. This allows for more precise control over the overall quality and diversity of the dataset. 
To implement this, we individually assess the quality and diversity of each sample and assign corresponding sampling weights based on these evaluations. Given a target token budget, we then sample each example according to its weight to construct the optimal training dataset.
Also, our approach offers the additional advantage of dynamically adapting to varying token budgets, enabling the determination of optimal data proportions for each specific budget. In contrast, the vast majority of existing works rely on static data proportions, which do not adjust to different token budget constraints.
The contributions of this paper are:
\begin{enumerate}
    \item We study the problem of sample-wise pre-training data mixing, which can alleviate the limitations of overlooking inter-domain overlap and suboptimal sample distribution within domains by existing domain-wise mixing works. 
    \item We propose a sample-wise pre-training data mixing strategy that coordinates data quality and diversity on a per-sample basis, effectively capturing commonalities among domains and optimal sample distribution.
    \item Extensive experiments on downstream tasks and perplexity evaluations demonstrate the advantages of our method. Notably, it achieves averaged baseline accuracy with 1.9x fewer training steps, highlighting its efficiency.
\end{enumerate}


\begin{figure*}[ht]
    \centering
    \includegraphics[width=0.9\linewidth]{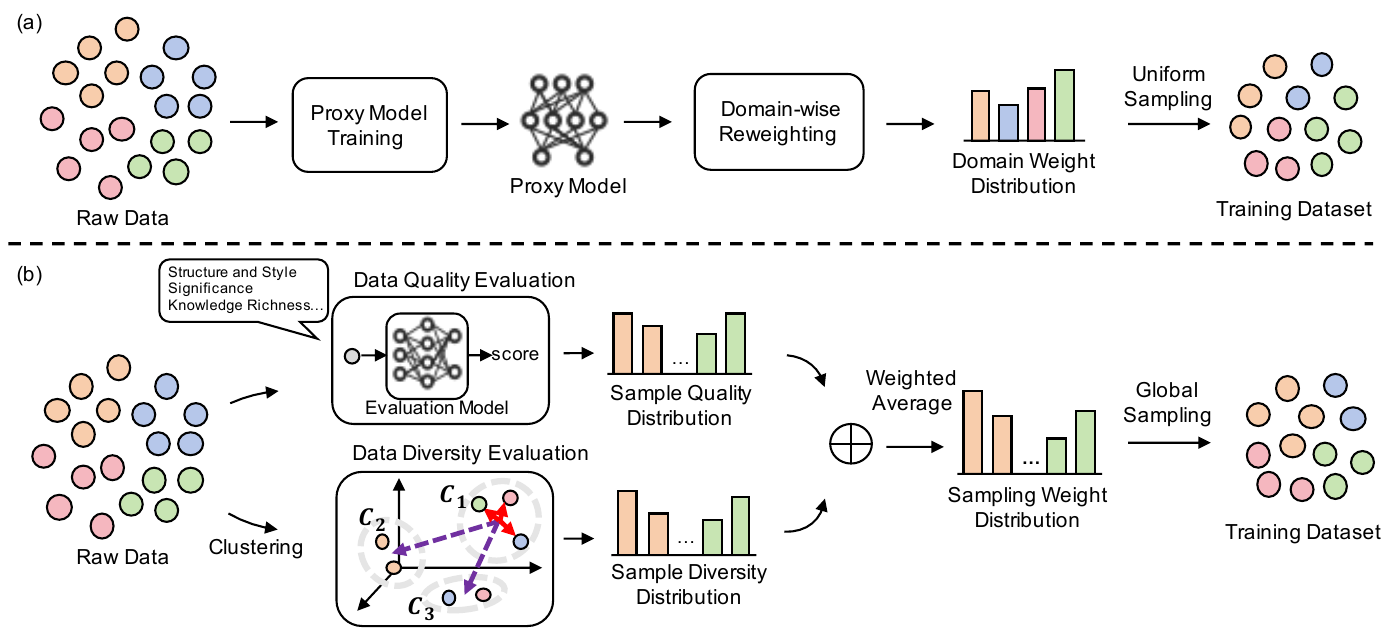}
    \caption{
        (a) Traditional methods determine domain weights and construct the training dataset by uniformly sampling from each domain.
        (b) \textit{SampleMix} employs a \textbf{sample-wise} mixing strategy by:
        evaluating sample quality and diversity,
        assigning appropriate weights, and
        constructing an optimal dataset based on these weights. Dots of the same color represent data from the same domain..
    }
    \label{fig:domainwise_vs_samplewise}
    \vspace{-4mm}
\end{figure*}

\section{Method}


\subsection{Problem Formulation}

Consider a source dataset $D_{\mathrm{src}}$ composed of $k$ distinct domains (e.g., CommonCrawl, Wikipedia, BookCorpus, etc.). For each domain $i$, let $D_i$ denote the collection of documents within that domain.
The entire source dataset is defined as $D_{\mathrm{src}} \triangleq \{D_1, \ldots, D_k\}$, with $T_{\mathrm{src}}$ representing the total number of tokens.
Our objective is to construct a target training set $D_{\mathrm{tgt}}$ for pre-training that adheres to a specific token budget $T_{\mathrm{tgt}}$ (e.g., 100B tokens). 
As illustrated in Figure \ref{fig:domainwise_vs_samplewise}, traditional approaches determine domain weights without explicitly considering the overall token budget, and build $D_{\mathrm{tgt}}$ by uniform sampling from each domain based on these weights.
In contrast, our proposed method, SampleMix, enhances this process by evaluating both the quality (\cref{section:quality}) and diversity (\cref{section:diversity}) of each document. Utilizing these dual criteria, SampleMix assigns unique sampling weights to each document. 
To ensure compliance with the token budget $T_{\mathrm{tgt}}$, we then construct an optimal training dataset by sampling documents according to their assigned weights (\cref{section:sampling}).


\subsection{Data Quality Evaluation}
\label{section:quality}

The quality of training data is crucial for large language models. However, most existing studies typically rely on simple heuristics \cite{xie2023data,li2023textbooks,sachdeva2024train}. 
\citet{wettig2024qurating} introduces four metrics and uses pairwise comparisons to train an evaluator model. However, these metrics are applied separately in data selection, and pairwise training may neglect the objective factors that determine sample quality.

\subsubsection{Quality Criteria}
To comprehensively capture both the fundamental linguistic attributes and the deeper informational and analytical qualities of the text, we assert that high-quality data should adhere to the following principles: linguistic precision and clarity, structural coherence and completeness, content reliability and appropriateness, informational and educational value, as well as significance and originality. To evaluate these aspects effectively, we propose 7 quality dimensions accompanied by corresponding scores based on the aforementioned principles, as outlined in Table \ref{tab:quality_dimensions}. 
Notably, for \textit{Knowledge Richness} and \textit{Logicality and Analytical Depth}, we utilize a larger scoring span \{0, 1, 2\} to address the wider range and greater complexity inherent in these features.
By aggregating all dimension scores, we obtain an overall quality evaluation for each sample, ranging from 0 to 10.

\begin{table}[!hbt]
    \centering
    \begin{tabular}{cc}
    \hline
    Dimension & Score \\
    \hline
     Clarity of Expression and Accuracy    &  \{0,1\}\\
     Completeness and Coherence    &  \{0,1\} \\
     Structure and Style & \{0,1\} \\
     Content Accuracy and Credibility & \{0,1\} \\
     Significance & \{0,1\} \\
     Knowledge Richness & \{0,1,2\} \\
     Logicality and Analytical Depth & \{0,1,2\} \\
    \hline
    \end{tabular}
    \caption{Quality dimensions and scores.}
    \label{tab:quality_dimensions}
    \vspace{-4mm}
\end{table}


\subsubsection{Quality Evaluator} 
To develop an effective and efficient quality evaluator, we utilize GPT-4o to assess training data based on predefined quality criteria (prompt shown in Fig \ref{fig:quality_prompt}). Specifically, we uniformly sample 420k documents from the SlimPajama dataset, allocating 410k and 10k documents for train and test set respectively. 
We train the quality evaluator with \texttt{gte-en-mlm-base} model \cite{zhang2024mgte} as the backbone.
Instead of text classification tasks, we employ ordinal regression to leverage the inherent ordering of quality scores. Following \citet{niu2016ordinal}, we transform ordinal regression into a series of binary classification problems, each indicating whether the input data exceeds a specific quality threshold. 
The overall quality score is then derived by subtracting the sequence of binary outputs (code shown in \cref{appendix:code_of_quality_evaluator}).

\begin{table}[!hbt]
    \centering
    \begin{tabular}{ccc}
    \hline
    Model & Text Classification  & Ordinal Regression \\
    \hline
    ACC & 56.14 &  55.94    \\
    MAE & 0.77 & 0.72 \\
    MSE & 1.95 & 1.57\\
    CACC &  82.24 &83.37 \\
    \hline
    \end{tabular}
    \caption{Performance comparison between text classification and ordinal regression models on the test set.}
    \label{tab:quality_evaluator_acc}
\end{table}

We evaluate the trained quality evaluator on the test set, as shown in Table \ref{tab:quality_evaluator_acc}. 
Instead of relying solely on Accuracy (ACC), we consider Mean Squared Error (MSE) and Mean Absolute Error (MAE), which more accurately reflect the degree of deviation between the true quality scores and the predicted results. 
While both the text classification and ordinal regression approaches achieve similar accuracy, the ordinal regression method demonstrates superior performance in terms of MSE and MAE.
We noticed that the accuracy is lower than anticipated; detailed analysis shows that most false predictions fall within ±1 of the true quality score. To address this, we introduce Close Accuracy (CACC), a relaxed metric where a prediction is considered correct if it is within ±1 of the true quality score. The CACC results indicate that our model possesses satisfactory discriminatory ability for samples of different qualities. 


\begin{figure}[ht]
  \centering
  \begin{subfigure}[b]{0.45\linewidth}
    \centering
    \includegraphics[width=\linewidth]{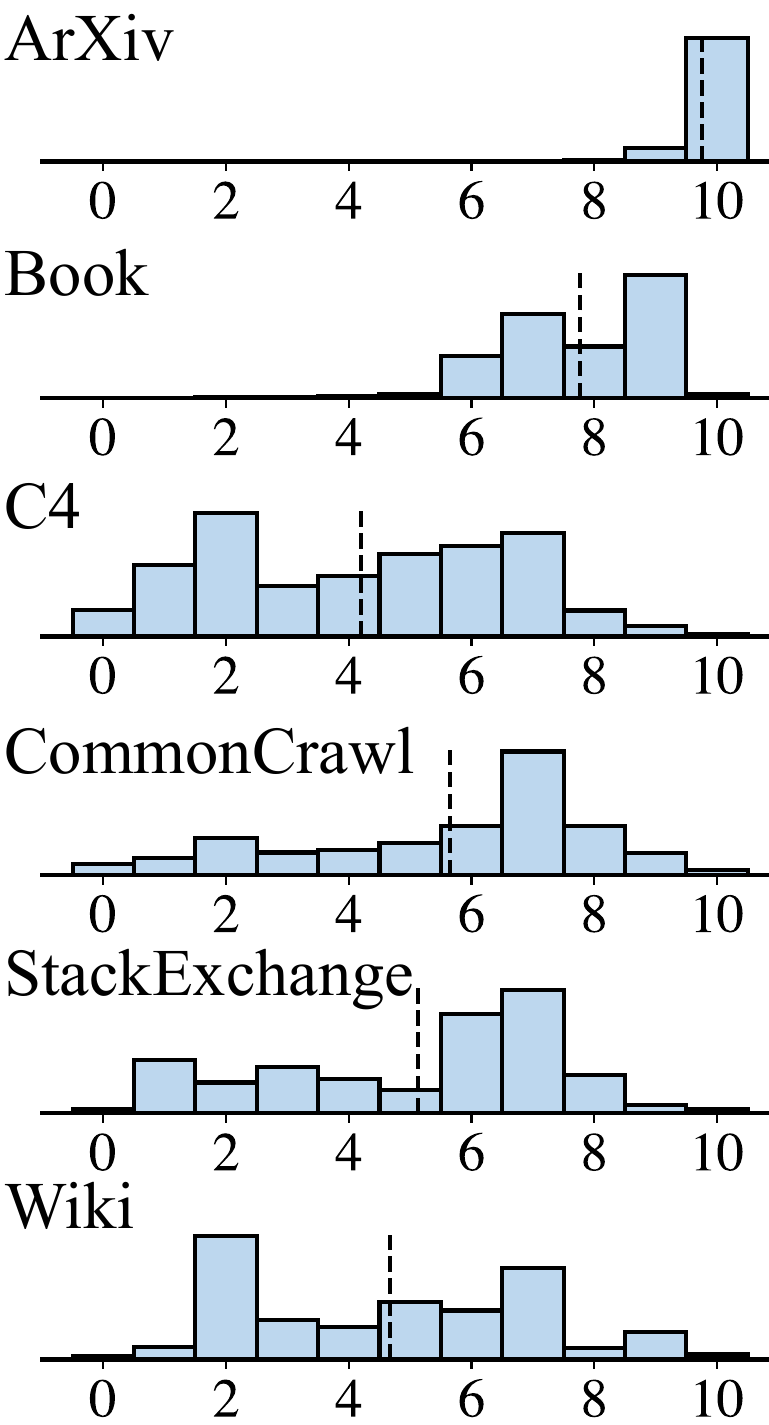}
    \caption{Quality Distribution}
    \label{fig:quality_distribution}
  \end{subfigure}
  \hfill
  \begin{subfigure}[b]{0.45\linewidth}
    \centering
    \includegraphics[width=\linewidth]{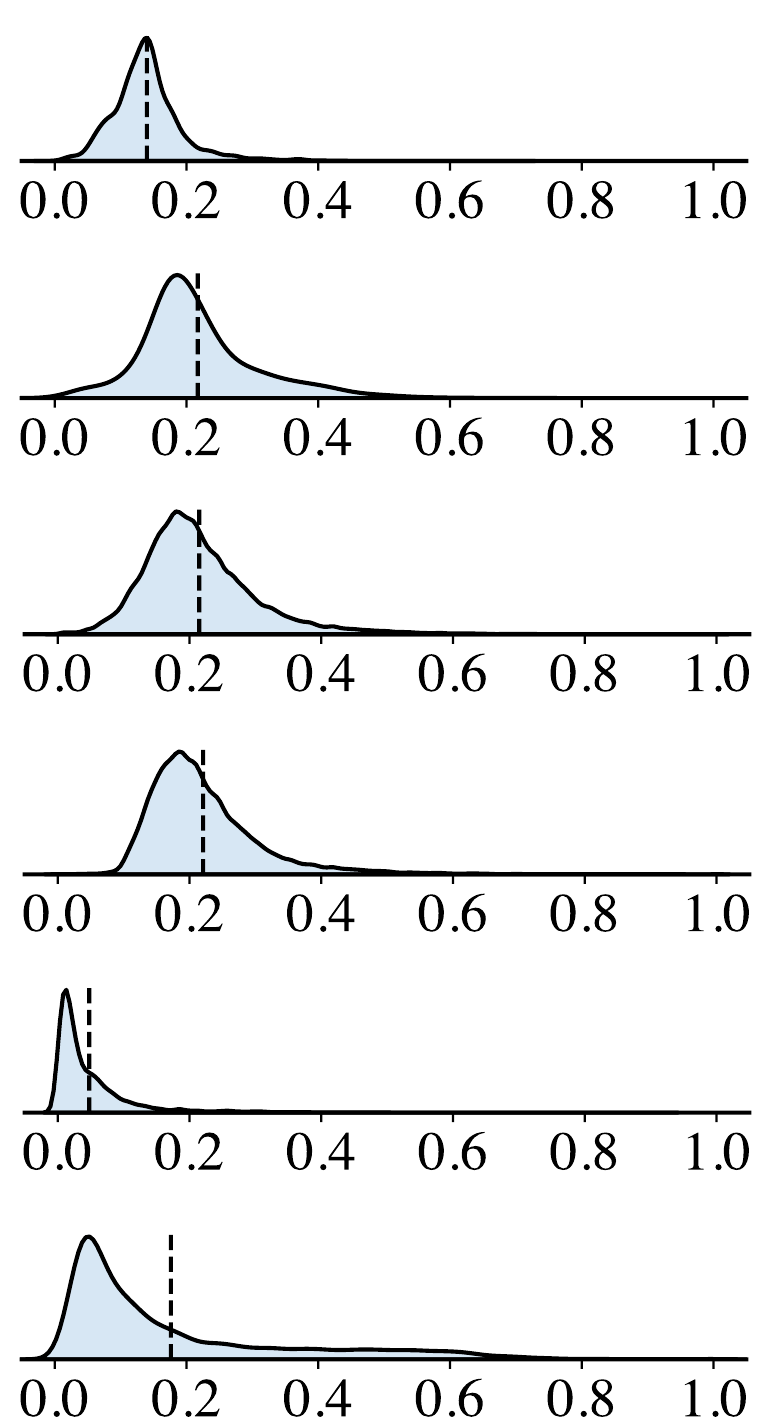}
    \caption{Diversity Distribution}
    \label{fig:diversity_distribution}
  \end{subfigure}
  \caption{Analysis of SlimPajama dataset. Mean values are marked with a dashed line.}
  \label{fig:distribution}
  \vspace{-4mm}
\end{figure}

\subsubsection{Analysis of Quality Distribution}
Using the trained quality evaluator, we annotated the SlimPajama dataset, and the resulting quality distribution is presented in Figure \ref{fig:quality_distribution}, from which we can find:
(1) Arxiv and Book sources exhibit higher quality, as anticipated.
(2) Wikipedia is generally considered a high-quality source; however, a substantial portion is of lower quality. Our manual inspection indicates that these low-quality samples typically consist of brief, parsing errors, incomplete content, and other issues.
(3) Overall, the CommonCrawl dataset outperforms C4 in terms of quality (average quality score: 5.65 v.s. 4.20).

\subsection{Data Diversity Evaluation}
\label{section:diversity}

Inspired by \citet{shao2024balanced} and \citet{abbas2024effective}, we employ data clustering to capture the text distribution within our training dataset. Through a detailed analysis of the clustered samples, we observe patterns consistent with \citet{abbas2024effective}'s work on image data, specifically:
(1) Denser clusters exhibit higher similarity among their constituent samples; 
(2) Clusters that are proximal to others are more likely to contain samples resembling those in neighboring clusters.
To quantify data diversity, we estimate a diversity measure for each sample using the Diversity Evaluator.

\subsubsection{Diversity Evaluator}

\textbf{Data Clustering} 
We begin by generating embeddings for each sample, which are subsequently organized into clusters via K-Means, effectively structuring the data based on textual similarity. The details of data clustering can be found in \cref{appendix:kmeans}.\\
\textbf{Cluster Compactness} 
We assess the density of a cluster by calculating the average distance of its members from the centroid, referred to as Cluster Compactness. A smaller average distance signifies a more compact cluster, indicating higher similarity among its constituent samples. This metric effectively reveals the dense property of the cluster.\\
\textbf{Cluster Separation} 
We evaluate the distinctiveness of each cluster by measuring the distance between its centroid and those of other clusters, termed Cluster Separation. Larger distances imply greater separation, indicating that the cluster is more distinct from others and highlighting its uniqueness on a global scale.\\
\textbf{Data Diversity Calculation} 
Finally, the diversity of each sample $x_i$ is estimated by integrating its cluster's separation and compactness as follows:
\begin{equation}
    d(x_i) =   d_{\text{compactness},j} \times d_{\text{separation},j}
\end{equation}
where \( x_i \) belongs to the \( j \)-th cluster, \( d_{\text{compactness},j} \) and \( d_{\text{separation},j} \) represents the cluster compactness and cluster separation for the \( j \)-th cluster respectively.
This composite diversity measure effectively encapsulates both the homogeneity within clusters and the distinctiveness between clusters, providing a comprehensive assessment of data diversity.

\subsubsection{Analysis of Diversity Distribution}


We examine the diversity distribution within the SlimPajama dataset, as illustrated in Figure \ref{fig:diversity_distribution}. We can find:
(1) Within individual domains, samples' diversity can vary significantly. For instance, the diversity distribution of C4 approximates a normal distribution, indicating consistent variability within this domain.
(2) Diversity differs markedly across domains in the SlimPajama dataset. Specifically, the C4, CommonCrawl, and Book domains exhibit the highest levels of diversity, as anticipated. In contrast, the StackExchange domain demonstrates the lowest diversity among the examined domains.



\subsection{Data Sampling}
\label{section:sampling}

\subsubsection{Sampling Weight Calculation}
Given the quality and diversity evaluation for each document, we first min-max normalize the dual measures to ensure they lie within the interval $[0,1]$ and compute the sampling weight as follows:
\begin{equation}
    p(x_i) = \alpha \, d(x_i) + (1 - \alpha) \, q(x_i)
\end{equation}
where $q(x_i)$ and $d(x_i)$ denote quality and diversity measure of the document $x_i$, and $\alpha \in [0,1]$ is the weighting factor that balances the contribution of diversity relative to quality.

\subsubsection{Determining Sampling Frequency}
Given the source dataset $D_{\mathrm{src}}$ containing $|D_{\mathrm{src}}|$ documents with $T_{\mathrm{src}}$ tokens, we first estimate the target number of documents for $D_{\mathrm{tgt}}$ as follows:
\begin{equation}
    |D_{\mathrm{tgt}}| = \frac{T_{\mathrm{tgt}}}{T_{\mathrm{src}}}|D_{\mathrm{src}}|
\end{equation}
Then we compute each document's sampling frequency $c(x_i)$ using a softmax-based distribution to translate the sampling weights into probabilities:
\begin{equation}
    c(x_i) = |D_{\mathrm{tgt}}| \times \frac{\exp\left(p(x_i)/\tau\right)}{\sum_{j \in D_{\mathrm{src}}} \exp\left(p(x_i)/\tau\right)}
\end{equation}
where $\tau$ is the temperature parameter that modulates the softmax distribution, controlling the concentration of the sampling probabilities.
\subsubsection{Constructing the Training Dataset}
Since $c(x_i)$ typically yields non-integer values, we convert these frequencies into integer counts through the following two-step process:
\begin{itemize}
    \item Integer Part: Always sample the document $\lfloor c(x_i) \rfloor$ times. For example, if $c(x_i) = 2.3$, the document is sampled 2 times.
    \item Fractional Part: The remaining fractional part ($c(x_i) - \lfloor c(x_i) \rfloor$) is used to determine an additional sample probabilistically. Continuing the example, with $c(x_i) = 2.3$,  there is a 30\% chance that $x_i$ will be sampled a third time, determined by comparing the fractional part to a randomly generated number.
\end{itemize}

By aggregating the sampled counts for each document $x_i$, we assemble the final training dataset $D_{\mathrm{tgt}}$, which closely matches the target token budget $T_{\mathrm{tgt}}$. 
Our method offers key benefits: 
(1) Prioritization of Quality and Diversity: 
By incorporating both quality and diversity metrics into the sampling weights, SampleMix ensures that high-quality and diverse documents are preferentially selected, enhancing the overall effectiveness of the training dataset.
(2) Adaptive to Training Budget: 
The sampling mechanism dynamically adjusts to different token budgets \( T_{\mathrm{tgt}} \), maintaining an optimal balance between quality and diversity without the need for manual tuning.
(3) Flexible Domain Representation:
By allowing different sampling rates within the same domain, the method supports a more nuanced representation of various domains.


\section{Experimental Setup}
\subsection{Dataset And Baselines}
\textbf{Dataset} Following \citet{xie2024doremi,ge2024data}, we experiment with the SlimPajama dataset, which consists of 7 domains from RedPajama, with intensive enhancements including NFC normalization, length filtering, and global deduplication \cite{cerebras2023slimpajama}.

\textbf{Baselines} We compare with the following baselines:
(1) \textbf{Vanilla}, which denotes the inherent proportions of datasets, mirroring the natural distribution patterns \cite{cerebras2023slimpajama}. 
(2) \textbf{DoReMi}, which exploits a learning-based solution for multi-round mixture optimization \cite{xie2024doremi}. 
(3) \textbf{CE}, which uses the Conditional Entropy proxy for data mixture optimization \cite{ge2024data}.  
(4) \textbf{BiMIX-OPT}, which derives the optimized data mixture by the bivariate scaling law \cite{ge2024data}. 
(5) \textbf{DoGE}, which determines the domain weight based on contribution to final generalization objective \cite{fan2023doge}.
(6) \textbf{DML}, which derives the optimized data mixture by the data mixing law \cite{ye2024data}. 
Note that we focus primarily on text data mixing. Following \citet{liu2024regmix}, we exclude the GitHub domain and apply re-normalization to the baseline weights (the weights are shown in Figure \ref{fig:mixtures}). Investigating code data mixing remains an avenue for future research.

\begin{table*}[!htb]
    \centering
    \begin{tabular}{lccccccc}
    \hline
    \textbf{Benchmark} & \textbf{Vanilla} & \textbf{DoReMi} & \textbf{CE} & \textbf{BiMIX-OPT} & \textbf{DoGE} & \textbf{DML} & \textbf{SampleMix} \\
    \hline
    \multicolumn{8}{c}{\emph{Downstream Tasks Evaluation (Accuracy)}} \\
    OpenBookQA      & 31.40 & 31.60 & \underline{31.80} & 29.80 & 29.00 & 30.80 & \textbf{32.60} \\
    LAMBADA         & 38.27 & \underline{40.95} & \textbf{42.23} & 38.02 & 37.07 & 35.40 & 40.69 \\
    PiQA            & 70.40 & 70.13 & 69.37 & 69.64 & \underline{70.62} & 65.02 & \textbf{70.95} \\
    ARC-Easy        & 47.44 & 46.65 & 46.73 & 45.57 & 45.74 & \underline{47.49} & \textbf{48.73} \\
    ARC-Challenge   & \underline{28.58} & 27.30 & 28.33 & 28.33 & 27.65 & 27.73 & \textbf{29.86} \\
    WinoGrande      & 52.33 & \textbf{54.38} & 51.07 & 52.80 & 51.14 & 51.46 & \underline{53.83} \\
    WiC             & 50.47 & 48.59 & 48.28 & 48.90 & 50.00 & \textbf{52.98} & \underline{51.72} \\
    RTE             & 50.18 & \underline{51.62} & \underline{51.62} & 47.65 & 51.26 & \underline{51.62} & \textbf{53.79} \\
    \hline
    \textbf{Average} & 46.13 & \underline{46.40} & 46.18 & 45.09 & 45.31 & 45.31 & \textbf{47.77} \\
    \hline
    \multicolumn{8}{c}{\emph{Perplexity Evaluation (Perplexity)}} \\
    Pile & 26.93 & 26.45 & \underline{26.20} & 27.47 & 29.49 & 29.76 & \textbf{25.63} \\
    xP3  & 47.38 & \underline{47.08} & 47.62 & 48.74 & 48.38 & 54.00 & \textbf{46.38} \\
    \hline
    \end{tabular}
    \caption{Comparison of data mixture methods across various downstream tasks and perplexity evaluations. The best performing method for each metric is highlighted in \textbf{bold}, while the second-best is \underline{underlined}.}
    \label{tab:main_results_acc}
    \vspace{-4mm}
\end{table*}

\subsection{Training Setup}
We train 1B-parameters LLaMA models \cite{dubey2024llama} from scratch with 100B tokens. 
For the baselines, we uniformly sample 100B tokens based on predefined domain weights.
Given that the source dataset (SlimPajama) comprises 503M documents totaling approximately 500B tokens, SampleMix generated the final training dataset consisting of 100M documents, with $\alpha$ and $\tau$ set to 0.8 and 0.2 respectively.
Detailed hyperparameters, including model architecture, learning rate, and other essential settings, are provided in Table \ref{tab:hyper-parameters}.


\subsection{Evaluation}
\textbf{Downstream Task Accuracy} Following \citet{fan2023doge,chen2025aioli}, we select 8 extensive downstream tasks, covering commonsense reasoning, language understanding, logical inference and general QA: OpenBookQA \cite{mihaylov2018can}, LAMBADA \cite{paperno2016lambada}, PiQA \cite{bisk2020piqa}, ARC-Easy, ARC-Challenge \cite{clark2018think}, WinoGrande \cite{sakaguchi2021winogrande}, and tasks from the SuperGLUE benchmark \cite{wang2019superglue}. We use LM-eval Harness \cite{eval-harness} and report the 5-shot accuracy.

\textbf{Validation Set Perplexity} Following \citet{ye2024data}, we compute perplexity on validation sets from \textit{The Pile} \cite{gao2020pile} to simulate separate collection of training and validation data. This metric measures the model's ability to predict text sequences accurately across various domains, reflecting its general language modeling proficiency.

\textbf{Instruction Tuning Perplexity} Following \citet{tirumala2023d4}, we evaluate perplexity on the instruction tuning dataset xP3 \cite{muennighoff2022crosslingual} to address the high variance in downstream tasks. This evaluation gauges the model's effectiveness in understanding and following instructions.




\section{Results and Analysis}
\subsection{Main Results}
Table \ref{tab:main_results_acc} presents the performance comparison between the baseline methods and our proposed SampleMix across downstream tasks and perplexity evaluations. We draw the following key observations:
(1) SampleMix achieves the highest average accuracy (47.77\%) across the eight downstream tasks, outperforming all baseline methods. Specifically, it leads in 5 out of 8 tasks, demonstrating its efficacy in enhancing performance.
(2) In perplexity evaluations, SampleMix records the lowest perplexity scores on both the Pile (25.63) and xP3 (46.38) datasets, underscoring the advantage in language modeling tasks.

\begin{figure}[!hbt]
    \centering
    \includegraphics[width=1.0\linewidth]{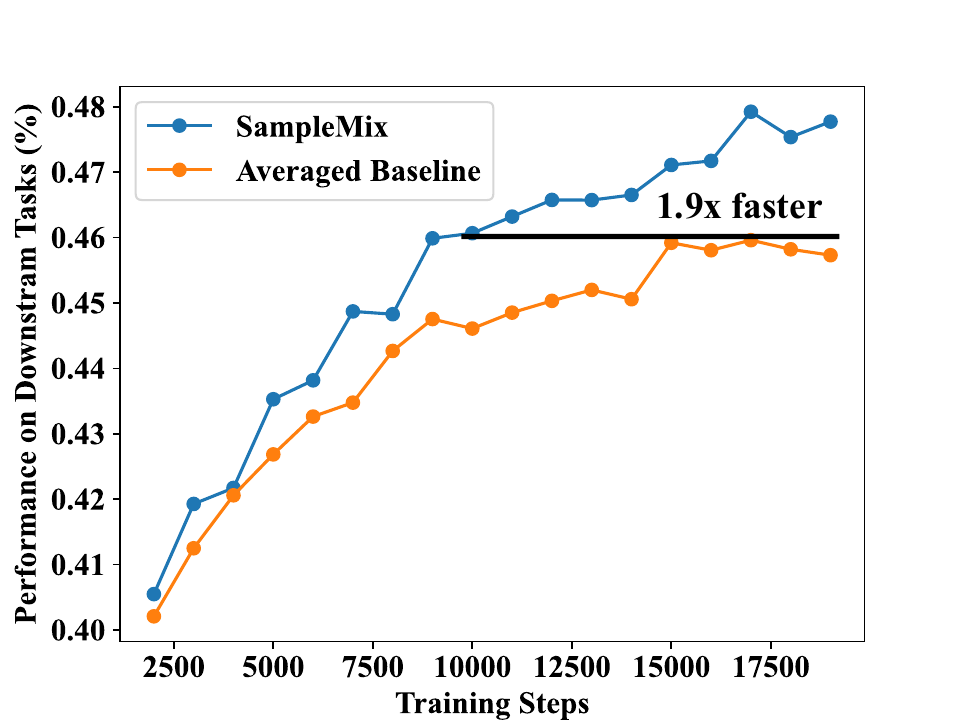}
    \caption{Training efficiency comparison. SampleMix reaches the average baseline accuracy at 100k training steps - 1.9 times faster than the averaged baselines.}
    \label{fig:speed_up}
    \vspace{-2mm}
\end{figure}

\textbf{Training Efficiency} 
We compare the convergence speed of SampleMix with that of baseline methods. SampleMix achieves the baselines' accuracy using 1.4x to 2.1x fewer training steps. As illustrated in Figure \ref{fig:speed_up}, it attains the average baseline accuracy within 100k steps—1.9x faster. This improvement demonstrates the efficiency gains provided by our approach. The full comparison is presented in Figure \ref{fig:speed_up_all}.


\textbf{Effectiveness on larger models} Furthermore, to assess the effectiveness on larger models, we trained 8B models using the top 3 performing baselines and SampleMix (training setup detailed in Table \ref{tab:hyper-parameters_8B}). As Table \ref{tab:8B_results} shows, SampleMix significantly outperforms the baselines, maintaining consistent advantages observed with 1B models.

\begin{table}[!hbt]
    \centering
    \begin{tabular}{cc}
    \hline
    Model     & Average Performance \\
    \hline
    Vanilla     &  53.17\\
    DoReMi & \underline{53.58}\\
    CE & 53.15 \\
    SampleMix & \textbf{54.86}\\
    \hline
    \end{tabular}
    \caption{Performance comparison with 8B models.}
    \label{tab:8B_results}
    \vspace{-2mm}
\end{table}

These results collectively demonstrate that SampleMix not only enhances overall performance but also does so with improved training efficiency. This establishes SampleMix as a robust and effective method for data mixture optimization.

\subsection{Effectiveness of Quality and Diversity}

\begin{figure}[!hbt]
    \centering
    \includegraphics[width=1.0\linewidth]{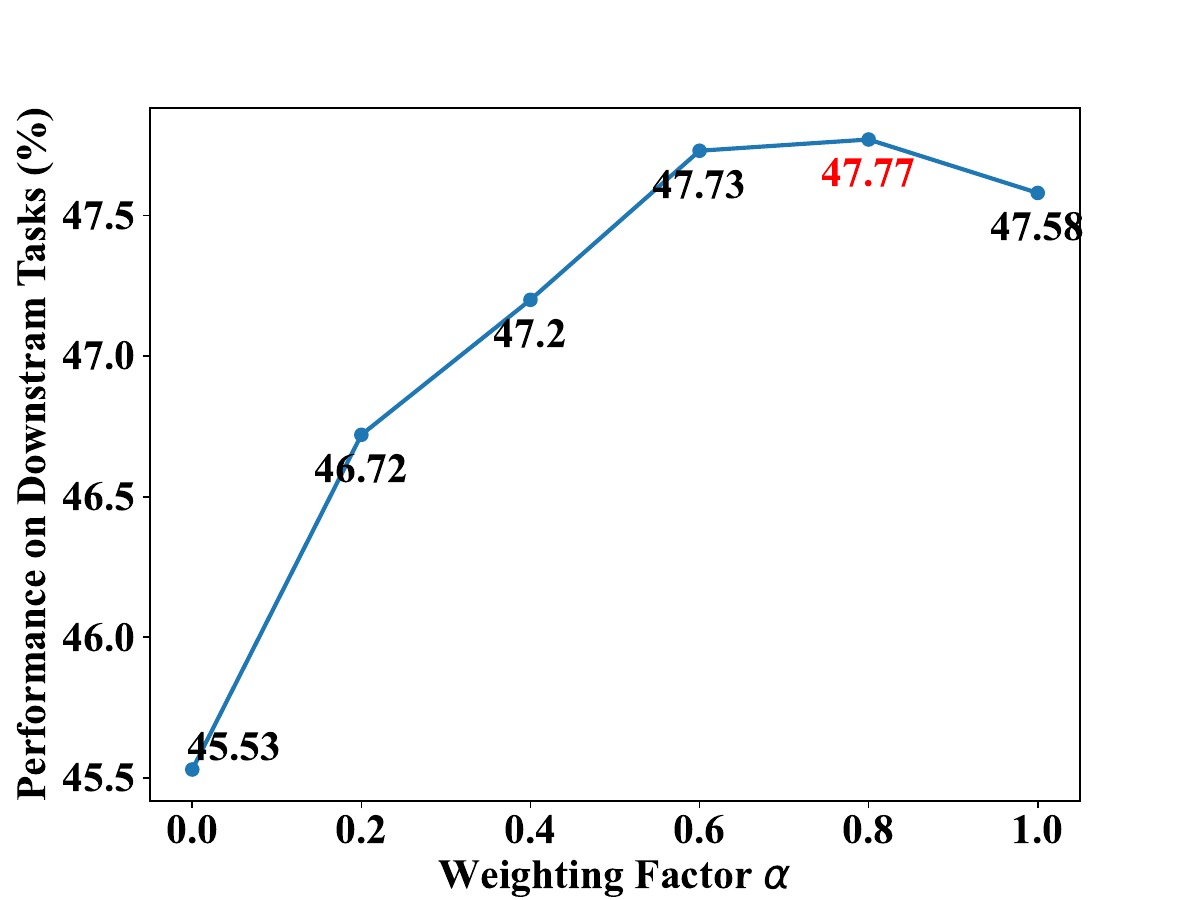}
    \caption{Average performance of downstream tasks with different weighting factor $\alpha$.}
    \label{fig:alpha_rsults}
    \vspace{-2mm}
\end{figure}

To further explore the effectiveness of our quality and diversity evaluation, we conducted a comprehensive analysis by systematically varying the weighting factor $\alpha$. Specifically, we performed a grid search with $\alpha$ values of 0.0, 0.2, 0.4, 0.6, 0.8, and 1.0. The corresponding model performances on downstream tasks are shown in Figure \ref{fig:alpha_rsults}.

From the results, we can observe the following:
\textbf{(1) Importance of Diversity} 
Setting $\alpha$ to 0.0 effectively excludes the diversity measure, relying solely on quality. This configuration yields the lowest accuracy of 45.53\%. As $\alpha$ increases from 0.0 to 0.8, there is a steady improvement in accuracy, peaking at 47.77\%.
This trend highlights the crucial role of diversity in achieving balanced data mixing and comprehensive data coverage.
\textbf{(2) Necessity of Quality} 
When $\alpha$ is set to 1.0, diversity is fully weighted, and quality is excluded, leading to a slight decrease in accuracy to 47.58\%. 
This minor drop indicates that while diversity is essential, incorporating the quality measure can further enhance performance.
\textbf{(3) Optimal Weighting} The optimal performance at $\alpha=0.8$ illustrates that prioritizing diversity while still valuing quality leads to the most effective model performance. We attribute the results to two factors. 
i) Measurement Scale: The absolute value of the diversity measure is inherently smaller compared to the quality measure. Consequently, even with a higher $\alpha$, the overall influence of diversity remains balanced when integrated with quality.
ii) Pre-processing Quality: The SlimPajama dataset has undergone rigorous quality filtering based on RedPajama, reducing the need for extensive weighting toward quality in the SampleMix framework.
\textbf{(4) Usage Recommendations} Users should adjust the weighting factor $\alpha$ based on the characteristics of their datasets. For example, in datasets with inherently lower quality, prioritizing quality yields better performance.

\subsection{Adaptation to Varying Token Budget}
Model development typically involves multiple training stages—such as pretraining, annealing, and continual pretraining—each requiring different token budgets. 
However, most existing methods present fixed data proportions, which limits their ability to accommodate varying token budget constraints effectively.
To evaluate the benefits of dynamically adapting to different token budgets, we scale the SlimPajama dataset to $\frac{1}{5}$ of its original size, resulting in a smaller source dataset ($\approx$ 100B tokens).
In our previous experiment, the full SlimPajama dataset served as the source dataset ($T_{\mathrm{src}}=500B$), while training was conducted with a subset of tokens ($T_{\mathrm{tgt}}=100B=\frac{1}{5}T_{\mathrm{src}}$). 
With the reduced source dataset, we adjusted the token budget proportion from $T_{\mathrm{tgt}}=\frac{1}{5}T_{\mathrm{src}}$ to $T_{\mathrm{tgt}}=T_{\mathrm{src}}$ (while maintaining $T_{\mathrm{tgt}}=100B$). We then conduct experiments under this adjusted token budget using the same setup.
As Table \ref{tab:main_results_acc_100B} shows, we can observe that:
(1) SampleMix achieves the highest average accuracy (47.46\%), demonstrating SampleMix's ability to effectively adapt to varying token budgets.
(2) Baseline methods exhibit inconsistent performance when the token budget changes. For instance, DoReMi, the best-performing baseline in previous experiments, underperforms Vanilla and CE. This inconsistency indicates that baseline methods struggle to adapt effectively to different token budgets.


\begin{table}[!hbt]
    \centering
    \begin{tabular}{cc}
    \hline
    Model & Average Performance \\
    \hline
    Vanilla    &  \underline{46.65}\\
    DoReMi     &  46.25\\
    CE & 46.40 \\
    BiMiX-OPT & 45.54\\
    DoGE & 45.01\\
    DML & 44.96\\
    SampleMix &  \textbf{47.46}\\
    \hline
    \end{tabular}
    \caption{Performance comparison of different data mixture methods with 100B data as candidate pool.}
    \label{tab:main_results_acc_100B}
    \vspace{-2mm}
\end{table}

\begin{figure}[!hbt]
  \centering
  \begin{subfigure}[b]{1.0\linewidth}
    \centering
    \includegraphics[width=\linewidth]{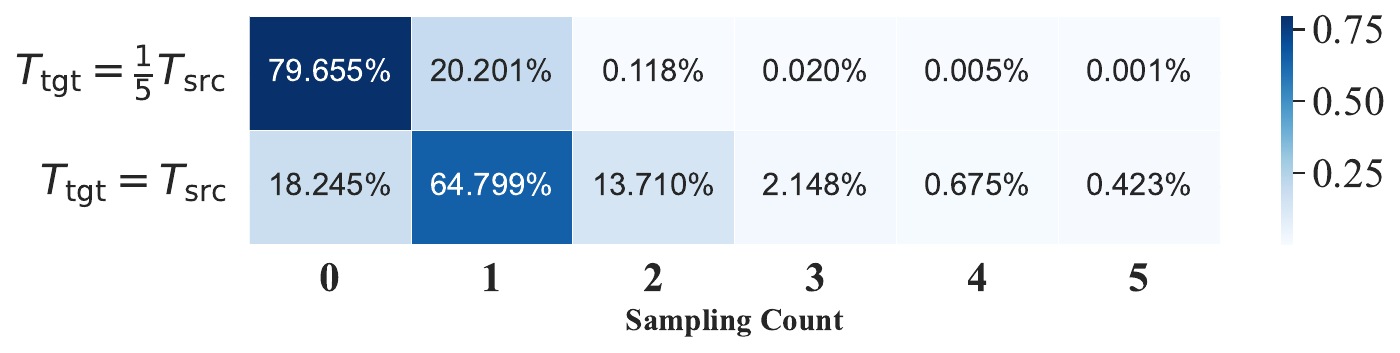}
    \caption{Proportion of different sampling counts.}
    \label{fig:average_sampling_count_prop}
  \end{subfigure}
  \hfill
  \begin{subfigure}[b]{1.0\linewidth}
    \centering
    \includegraphics[width=\linewidth]{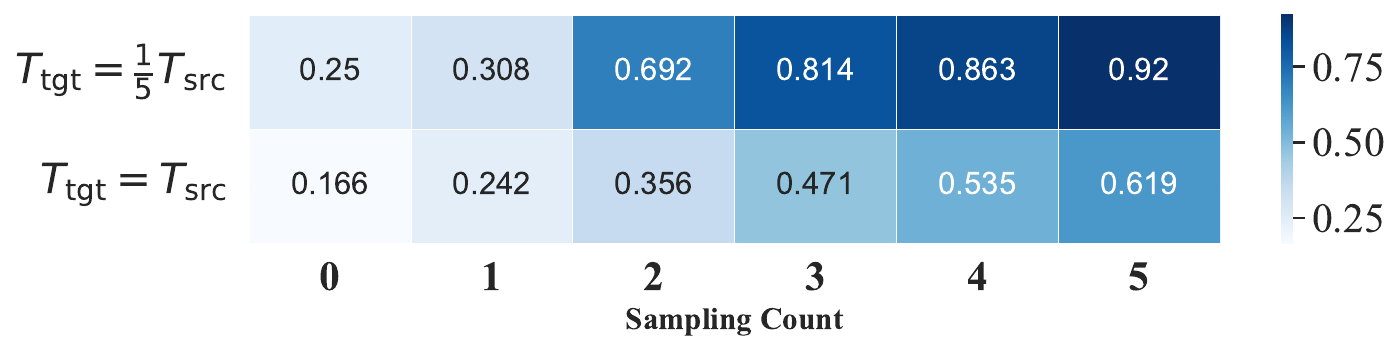}
    \caption{Sampling weight (i.e., $p(x)$) of different sampling counts.}
    \label{fig:average_sampling_count_px}
  \end{subfigure}
  \caption{Analysis of different sampling counts.}
  \label{fig:average_sampling_count}
  \vspace{-4mm}
\end{figure}

To further investigate how SampleMix adapts to varying token budgets, we analyze the sampling counts under different scenarios. Figure \ref{fig:average_sampling_count_prop} illustrates the proportion of various sampling counts, while Figure \ref{fig:average_sampling_count_px} presents the average sampling weights $p(x)$ associated with these counts.
We can observe that:
For $T_{\mathrm{tgt}}=\frac{1}{5}T_{\mathrm{src}}$, the source dataset is sufficiently large, allowing top-tier data to meet the token budget.
SampleMix precisely selects high-weight samples to fulfill the budget requirements, minimizing the need for extensive upsampling (i.e., sampling count > 1 is rare) and ensuring that all valuable data is included.
For $T_{\mathrm{tgt}}=T_{\mathrm{src}}$, the source dataset is relatively smaller, and high-weight samples alone are insufficient to meet the token budget. To satisfy the budget, SampleMix incorporates lower-weight samples. Despite this inclusion, the method effectively identifies and discards the least valuable data, which accounts for 18.245\% of the dataset due to their low sampling weights (average weight = 0.166). Data with higher sampling weights are upsampled more frequently, thereby enhancing their representation within the constrained budget.
Additionally, for $T_{\mathrm{tgt}}=\frac{1}{5}T_{\mathrm{src}}$, the average sampling weight is larger (0.312 v.s. 0.289 when $T_{\mathrm{tgt}} = T_{\mathrm{src}}$), 
further verifying SampleMix's ability to effectively utilize the sampling space and adapt to varying token budgets.

\section{Related Work}
We have covered research on data mixture in \cref{sec:introduction}, related work related to our technical designs is mainly introduced in the following.

\textbf{Data Quality}
Heuristic rules, such as thresholds on word repetitions and perplexity, are commonly used to filter out low-quality data \cite{YUAN202165,dodge2021documenting,laurenccon2022bigscience} . 
Earlier model-based methods employ binary classifiers to distinguish high-quality from low-quality data \cite{brown2020language}.
Recent approaches incorporated more sophisticated models. 
\citet{sachdeva2024train} proposes the ASK-LLM sampler, which evaluates data quality by asking for a proxy LLM. 
\citet{wettig2024qurating} investigated four qualities-writing style, required expertise, facts \& trivia, and educational value respectively.
However, most methods rely on relatively coarse criteria and do not fully leverage the multi-dimensional property of data quality.


\textbf{Diversity}
Traditional deduplication methods struggle to capture more complex semantic similarities \cite{wenzek2020ccnet,soldaini-etal-2024-dolma}.
To better handle semantic redundancy, \citet{abbas2023semdedup} applies K-Means clustering in the embedding space to identify and remove redundant data. \citet{tirumala2023d4} builds on this approach by using SemDeDup as a preprocessing step before applying SSL Prototypes \cite{sorscher2022beyond}.
\citet{shao-etal-2024-balanced} balances common and rare samples and ensures diversity by data clustering.
\section{Conclusion}
We have presented SampleMix, a sample-wise pre-training data mixing strategy by coordinating data quality and diversity. Extensive experiments demonstrate that SampleMix outperforms existing domain-wise methods, achieving comparable accuracy with 1.9x fewer training steps. 
In the future, we are interested in incorporating automatic evaluation metrics derived from the model's perspective to complement the current manually designed measures, and exploring code data mixing.



\newpage
\section{Limitations}

In this study, we conducted experiments exclusively using the SlimPajama dataset and identified the optimal hyperparameters specific to this dataset. Consequently, the hyperparameter settings reported may not directly transfer to other datasets with different characteristics. Users aiming to apply our methodology to their own datasets will need to perform tailored hyperparameter tuning to achieve optimal performance.
Specifically, we suggest assigning a smaller $\alpha$
 to prioritize data quality in lower-quality datasets, thereby minimizing the influence of subpar data. Conversely, for higher-quality datasets, a larger $\alpha$ is recommended to ensure comprehensive data coverage through increased diversity. However, the optimal balance between diversity and quality may vary depending on the specific attributes and complexities of different datasets.

\bibliography{acl_latex}
\newpage
\newpage
\appendix

\section{Domain Overlaps}
We manually check the samples within the same cluster but from different domains. Such samples are usually topic-relevant and similar in terms of structure, semantics, and context. As Figure \ref{fig:samples_of_same_topic} shows, the samples all discuss topics about Einstein and the Theory of Relativity.

\label{appendix:domain_overlaps}
\begin{figure*}[!hbt]
\centering
\begin{tcolorbox}[
    colback=gray!10, 
    colframe=gray, 
    title=Samples from Different Data Sources with Similar Topics,
    width=\textwidth, 
    fonttitle=\bfseries,
    boxrule=0.5pt,            
    boxsep=3.5pt, 
    left=5pt, 
    right=5pt, 
    top=5pt, 
    bottom=5pt
]
    \small 
    \begin{tabular}{p{0.48\textwidth} p{0.48\textwidth}}
        \textbf{Arxiv} & \textbf{C4} \\
        General relativity suffers from a number of problems regarding its local conservation laws for energy and momentum. This was the subject of a crucial discussion between Hilbert, Klein, Noether, and Einstein between 1915 and 1918.... &
        The term $mc^2$ had already made an appearance in his paper of 26 September, which introduced special relativity. The paper of 21 November showed that $E=mc^2$ applies to bodies at rest. [Physics Today]... \\
        \smallskip & \smallskip \\
        \textbf{CommonCrawl} & \textbf{StackExchange} \\
        The General Theory of Relativity (GRT) was born among other things from the demand to be able to use arbitrary coordinate systems for the description of the laws of nature. According to the covariance principle, the form of the laws of nature should not depend decisively on the choice of the special coordinate system... &
        No one but Einstein can be sure of exactly how he arrived at GR. From reading various histories of the time it seems to me that once Einstein had come up with the equivalence principle he started looking around for theories that embodied it... \\
        \smallskip & \smallskip \\
        \multicolumn{2}{p{0.96\textwidth}}{
            \textbf{Wikipedia} \newline
            \emph{The Meaning of Relativity: Four Lectures Delivered at Princeton University, May 1921} is a book published by Princeton University Press in 1922 that compiled the 1921 Stafford Little Lectures at Princeton University, given by Albert Einstein...
        } \\
    \end{tabular}
\end{tcolorbox}
\caption{Samples from different domains, all describing information related to Einstein and Theory of Relativity.}
\label{fig:samples_of_same_topic}
\end{figure*}

\section{Samples from Slimpajama CommonCarwl}
We manually check the low-quality and high-quality samples from Slimpajama CommonCarwl. As Figure \ref{fig:samples_of_cc} shows, the data quality of CommonCrawl varies significantly. The low-quality sample is characterized by fragmented and disorganized information, primarily consisting of sporadic headlines and links related to sports news. On the other hand, the high-quality sample provides a coherent and informative excerpt about astrophysical research, demonstrating a clear and structured narrative.
\begin{figure*}[ht]
\begin{tcolorbox}[
    colback=gray!10,          
    colframe=gray,            
    title=Low-Quality and High-Quality Samples, 
    boxrule=0.5pt,            
    boxsep=3.5pt,
    fonttitle=\bfseries,      
    left=5pt, right=5pt,      
    top=5pt, bottom=5pt       
]
  \textbf{Low-Quality Sample} \\
New posts Featured Search forums

Sports Briefing (New York Times)

Thread starter articlebot

Cycling News Headlines

articlebot

auto racing.

http://us.rd.yahoo.com/dailynews/rss/search/cycling+racing/SIG=120pnaegk/*http

Cycling News Headlines Jul 31, 2007

Cycling News Headlines Jun 2, 2007

Cycling News Headlines May 11, 2007

Cycling News Headlines Mar 17, 2007

Cycling News Headlines Dec 20, 2006

Sports Briefing: Basketball, Cycling, Auto Racing, Hockey, Golf, Football and Soccer (New York Times

Cycling News Headlines Nov 26, 2006

Cycling News Headlines Oct 17, 2006

Cycling News Headlines Sep 21, 2006

Sports Briefing: Track and Field, Marathon, Auto Racing, College Football and Cycling (New York Time

Cycling News Headlines Aug 28, 2006

Sports Briefing: Baseball, Golf, Horse Racing and Cycling (New York Times)

Cycling News Headlines Jun 26, 2006
  \tcblower
  \textbf{High-Quality Sample} \\
Decades of studies show that most massive galaxies harbor a supermassive black hole at their center, with the mass of the black hole being one tenth of the total mass of the surrounding spheroid of stars. Two astrophysicists from the Center for Astrophysics | Harvard and the Smithsonian have proposed a method to observe what could be the second-closest supermassive black hole to Earth. 
\end{tcolorbox}
\caption{Quality of CommonCrawl Samples may vary significantly.}
\label{fig:samples_of_cc}
\end{figure*}

\section{Domain Weights of Different Methods}

\begin{figure*}
    \centering
    \includegraphics[width=0.8\linewidth]{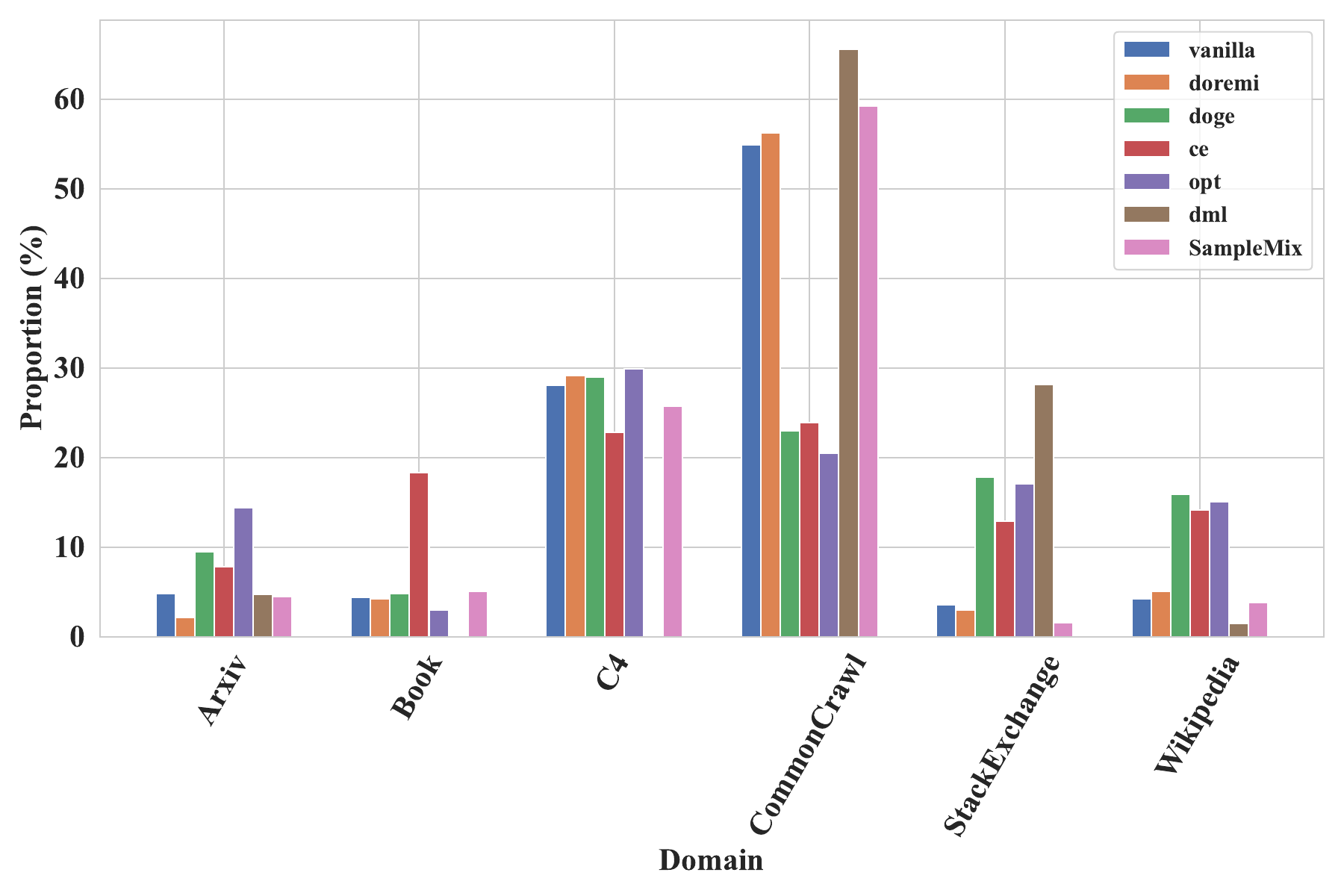}
    \caption{Domain weights of different methods.}
    \label{fig:mixtures}
\end{figure*}

Figure \ref{fig:mixtures} shows the domain weights of different methods.

\section{Hyper-Parameters of Training Models}
The experiments for both 1B and 8B parameter models follow standard transformer architecture with carefully optimized hyper-parameters. Table \ref{tab:hyper-parameters} and Table \ref{tab:hyper-parameters_8B} introduce the architectural configurations and training specifications for both model scales respectively.
\begin{table}[!hbt]
    \centering
    \begin{tabular}{cc}
    \hline
    Hyper-parameter & Value \\
    \hline
    layer num & 28\\
    attention head num     &  13 \\
    attention head dim & 128 \\
    model dim & 1664\\
    ffn intermediate dim & 4480\\
    global batch size & 1280 \\
    sequence length & 4096 \\
    learning rate & $2e^{-4}$ \\
    learning rate scheduler & cosine scheduler \\
    learning rate warmup tokens & 525M \\
    \hline
    \end{tabular}
    \caption{Hyper-parameters of 1B models used in the experiment.}
    \label{tab:hyper-parameters}
\end{table}

\begin{table}[!hbt]
    \centering
    \begin{tabular}{cc}
    \hline
    Hyper-parameter & Value \\
    \hline
    layer num & 32\\
    attention head num     &  32 \\
    attention head dim & 128 \\
    model dim & 4096\\
    ffn intermediate dim & 14336\\
    global batch size & 1280 \\
    sequence length & 4096 \\
    learning rate & $2e^{-4}$ \\
    learning rate scheduler & cosine scheduler \\
    learning rate warmup tokens & 525M \\
    \hline
    \end{tabular}
    \caption{Hyper-parameters of 8B models used in the experiment.}
    \label{tab:hyper-parameters_8B}
\end{table}

\section{Quality Evaluation Prompt}
\label{appendix:quality_evaluation_prompt}
The prompt for GPT-4o to assess training data quality is given in Figure \ref{fig:quality_prompt}.
\begin{figure*}[ht]
    \begin{tcolorbox}[
    colback=gray!10,          
    colframe=gray,            
    title=Quality Evaluation Template, 
    boxrule=0.5pt,            
    boxsep=3.5pt,
    fonttitle=\bfseries, 
    left=5pt, right=5pt,      
    top=5pt, bottom=5pt,      
]
\small
Annotator Task: Text Data Quality Evaluation

Role: You are a Language Model Training Data Annotator. Your job is to evaluate the quality of text documents.

Objective: Assess each document using the seven evaluation dimensions below. For each dimension, assign a score based on the provided criteria to determine the document's quality.

Evaluation Dimensions:

1. Clarity of Expression and Accuracy (0-1 points)

- Evaluate: How clearly ideas are expressed and the correctness of language (grammar, syntax, punctuation).

- Score:

 \ - 0: Numerous grammatical, spelling, or punctuation errors that significantly hinder comprehension.

 \ - 1: Few grammatical or punctuation errors that do not impede understanding; ideas are clearly and smoothly expressed.

2. Completeness and Coherence (0-1 points)

- Evaluate: Whether paragraphs are fully developed, relevant to the main theme, and logically connected.

- Score:

 \ - 0: Underdeveloped or off-topic paragraphs; lack of logical flow causing confusion.

 \ - 1: Well-developed, relevant paragraphs that are logically connected and contribute to a unified theme.

3. Structure and Style (0-1 points)

- Evaluate: The overall logical flow of the document and the clarity of the author's presentation.

- Score:

 \ - 0: Unclear structure and inconsistent or unengaging style.

 \ - 1: Clear and logical structure with a consistent and appropriate style that facilitates understanding.

4. Content Accuracy and Credibility (0-1 points)

- Evaluate: Appropriateness of content (free from pornography, drugs, violence) and the accuracy and reliability of facts and sources.

- Score:

 \ - 0: Inappropriate material or contains factual inaccuracies and unreliable sources.

 \ - 1: Appropriate content, free from prohibited material, with accurate and credible information.

5. Significance (0-1 points)

- Evaluate: The importance, originality, and broader impact of the document compared to others in the field. Verify that the document is not machine-generated.

- Score:

 \ - 0: Lacks importance and originality. It does not provide unique insights or contribute meaningfully beyond its immediate purpose. It is not recognized as historically significant or exhibits characteristics of being machine-generated.

 \ - 1: Demonstrates originality and important or impactful. It demonstrates originality and offers unique insights or contributions. It may also hold historical significance or be recognized as influential.

6. Knowledge Richness (0-2 points)

- Evaluate: The depth and breadth of information, including comprehensive insights and detailed explanations that enhance the reader's understanding. Ensure that any concepts or jargon used are well-explained.

- Score:

 \ - 0: Minimal information with little to no depth or insights.

 \ - 1: Adequate information with some insightful explanations; concepts or jargon introduced but not thoroughly explained.

 \ - 2: Comprehensive and detailed information with deep insights; all concepts and jargon are clearly explained and accessible, offering strong educational value.

7. Logicality and Analytical Depth (0-3 points)

- Evaluate: The text's ability to present profound insights or viewpoints, supported by in-depth analysis and reasoning.

- Score:

 \ - 0: Contains only simple statements and basic facts without deeper exploration.

 \ - 1: Describes or analyzes straightforward issues or processes with limited depth.

 \ - 2: Offers detailed analysis or solutions, addressing complex professional issues with substantial depth.

 \ - 3: Building on the 2-point criteria, if the text involves STEM fields (Science, Technology, Engineering, Mathematics), such as astronomy, medicine, mathematics, physics, chemistry, biology, etc., an additional point is awarded for a total of 3 points, acknowledging the specialized complexity and depth required in these areas.

Requirements: Based on the above dimensions, score the text content, first stating the evaluation reasons, then providing the quality assessment score. The final score is the sum of all dimensions, ranging from 0-10 points. Output format is JSON:

\{"Evaluation Reasons": {"Clarity of Expression": "...", "Completeness and Coherence": "...", "Structure and Style": "...", "Appropriate Content and Credibility": "...", "Significance": "...", "Knowledge Richness": "...", "Logicality and Analytical Depth": "..."}, "Clarity of Expression": X, "Completeness and Coherence": X, "Structure and Style": X, "Appropriate Content and Credibility": X, "Significance": X, "Knowledge Richness and Educational Value": X, "Logicality and Analytical Depth": X, "Final Score": X\}

Evaluate all the text as a whole:

<<<Document>>>
\end{tcolorbox}
\caption{Prompt for GPT-4o to assess training data quality.}
\label{fig:quality_prompt}
\end{figure*}

\section{Code of Quality Evaluator}
Table \ref{tab:ordinal_model} shows the Python code for implementing the ordinal regression model aimed at quality scoring tasks, including model definition, loss function computation, and inference process. The full
code can be found in the supplementary materials.

The \texttt{OrdinalRegressionModel} class initializes the pre-trained base model and a series of ordinal layers. Each ordinal layer outputs the probability that the quality score is greater than a specific threshold. For instance, the first ordinal layer (index 0) computes the probability that the quality score is greater than 0, i.e., the probability that the score is at least 1. Similarly, the second ordinal layer (index 1) calculates the probability that the quality score is greater than 1, meaning the probability that the score is at least 2, and so on. The last ordinal layer (index 9) computes the probability that the score is greater than 9, which is equivalent to the probability that the score is exactly 10. Therefore, the model has 10 ordinal layers in total, each corresponding to one of these thresholds. 

The \texttt{loss} function calculates the ordinal loss by summing the binary cross-entropy loss between the predicted probabilities and the target values. For each ordinal layer, a binary target is created, indicating whether the true score is greater than the threshold corresponding to that layer. Specifically, the larger the deviation between the predicted score and the true score, the higher the loss, which helps the model focus on reducing these deviations during training.

The \texttt{predict} function implements inference using the trained ordinal regression model. It first computes the predicted probabilities for each class, and then calculates the final predicted score by selecting the class with the maximum probability. The function also calculates the probability distribution across all possible scores, which provides a measure of confidence for the predicted score.
\label{appendix:code_of_quality_evaluator}
\begin{table*}[ht]
\centering
\lstset{
    basicstyle=\ttfamily\footnotesize, 
    numbers=left, 
    numberstyle=\tiny\color{gray}, 
    keywordstyle=\color{blue}, 
    commentstyle=\color{green!50!black}, 
    stringstyle=\color{red}, 
    backgroundcolor=\color{gray!10}, 
    frame=tb, 
    rulecolor=\color{black}, 
    tabsize=4, 
    captionpos=b, 
    breaklines=true, 
    breakatwhitespace=true, 
    showspaces=false, 
    showstringspaces=false, 
    showtabs=false, 
    morekeywords={*,...} 
}
\begin{lstlisting}[language=Python]
# Define the ordinal regression model class
class OrdinalRegressionModel(torch.nn.Module):
    def __init__(self, pretrained_path, num_classes=10):
        super(OrdinalRegressionModel, self).__init__()
        self.base_model = AutoModel.from_pretrained(pretrained_path)
        self.ordinal_layers = torch.nn.ModuleList([torch.nn.Linear(
                              self.base_model.config.hidden_size, 1)
                              for _ in range(num_classes)])
    
    def forward(self, input_ids, attention_mask=None, token_type_ids=None):
        outputs = self.base_model(input_ids=input_ids,
                                  attention_mask=attention_mask,
                                  token_type_ids=token_type_ids)
        last_hidden_state = outputs.last_hidden_state
        cls_representation = last_hidden_state[:, 0, :]
        
        # Compute the output for each ordinal layer
        ordinal_outputs = [torch.sigmoid(layer(cls_representation))
                           for layer in self.ordinal_layers]
        ordinal_outputs = torch.cat(ordinal_outputs, dim=1)
        return ordinal_outputs

# Calculate the ordinal loss
def loss(outputs, targets):
    loss = 0.0
    for i in range(outputs.size(1)):
        binary_targets = (targets > i).float()
        loss += nn.functional.binary_cross_entropy(outputs[:, i], binary_targets)
    return loss

# Inference function
def predict(text):
    with torch.no_grad():
        inputs = tokenizer(
            text,
            truncation=True,
            padding=True,
            max_length=4096,
            return_tensors="pt"
        )
        # Get model outputs
        outputs = model(input_ids=inputs['input_ids'],
                        attention_mask=inputs['attention_mask'])
        
        # Initialize probability array
        probabilities = torch.zeros(outputs.size(0), outputs.size(1) + 1)
        # Calculate probability for the first class
        probabilities[:, 0] = 1 - outputs[:, 0]  
        if outputs.size(1) > 1:
            # Calculate probabilities for the middle classes
            probabilities[:, 1:-1] = outputs[:, :-1] - outputs[:, 1:]
        # Calculate probability for the last class
        probabilities[:, -1] = outputs[:, -1]  

        # Calculate scores by finding the index of the maximum probability
        scores = torch.argmax(probabilities, dim=1)
    return scores, probabilities
\end{lstlisting}
\caption{Python Code for implementing the ordinal regression model.}
\label{tab:ordinal_model}
\end{table*}

\section{K-means Clustering Details}
\label{appendix:kmeans}
For the data clustering in \cref{section:diversity}, we generate 768-dimensional embeddings for each sample \footnote{https://huggingface.co/princeton-nlp/unsup-simcse-bert-base-uncased}. Further, we normalize the embeddings to have L2-norm of 1.0, and use faiss \cite{johnson2019billion} to perform K-means clustering.
Following \citet{tirumala2023d4,abbas2024effective}, we set the number of clusters to be the square root of the number of total points being clustered. The core code of data clustering is presented in Table \ref{tab:kmeans_clustering}. The full code can be found in the supplementary materials.

\begin{table*}[ht]
\centering
\lstset{
    basicstyle=\ttfamily\footnotesize, 
    numbers=left, 
    numberstyle=\tiny\color{gray}, 
    keywordstyle=\color{blue}, 
    commentstyle=\color{green!50!black}, 
    stringstyle=\color{red}, 
    backgroundcolor=\color{gray!10}, 
    frame=tb, 
    rulecolor=\color{black}, 
    tabsize=4, 
    captionpos=b, 
    breaklines=true, 
    breakatwhitespace=true, 
    showspaces=false, 
    showstringspaces=false, 
    showtabs=false, 
    morekeywords={*,...} 
}
\begin{lstlisting}[language=Python]
    # Calculate the number of clusters
    n_centroids = int(math.sqrt(all_embeddings.shape[0]))
    # define the parameters
    kmeans = faiss.Kmeans(
        d = 768, 
        k = n_centroids, 
        niter=50, # 50 iterations
        gpu = True, 
        seed = 1024, 
        spherical = True, 
        min_points_per_centroid=1,
        max_points_per_centroid=all_embeddings.shape[0]
    )
    # perform data clustering
    kmeans.train(all_embeddings)
\end{lstlisting}
\caption{Python Code for implementing K-Means clustering.}
\label{tab:kmeans_clustering}
\end{table*}

\section{Coverage Speed of All Methods}
\begin{figure*}[!hbt]
    \centering
    \includegraphics[width=0.9\linewidth]{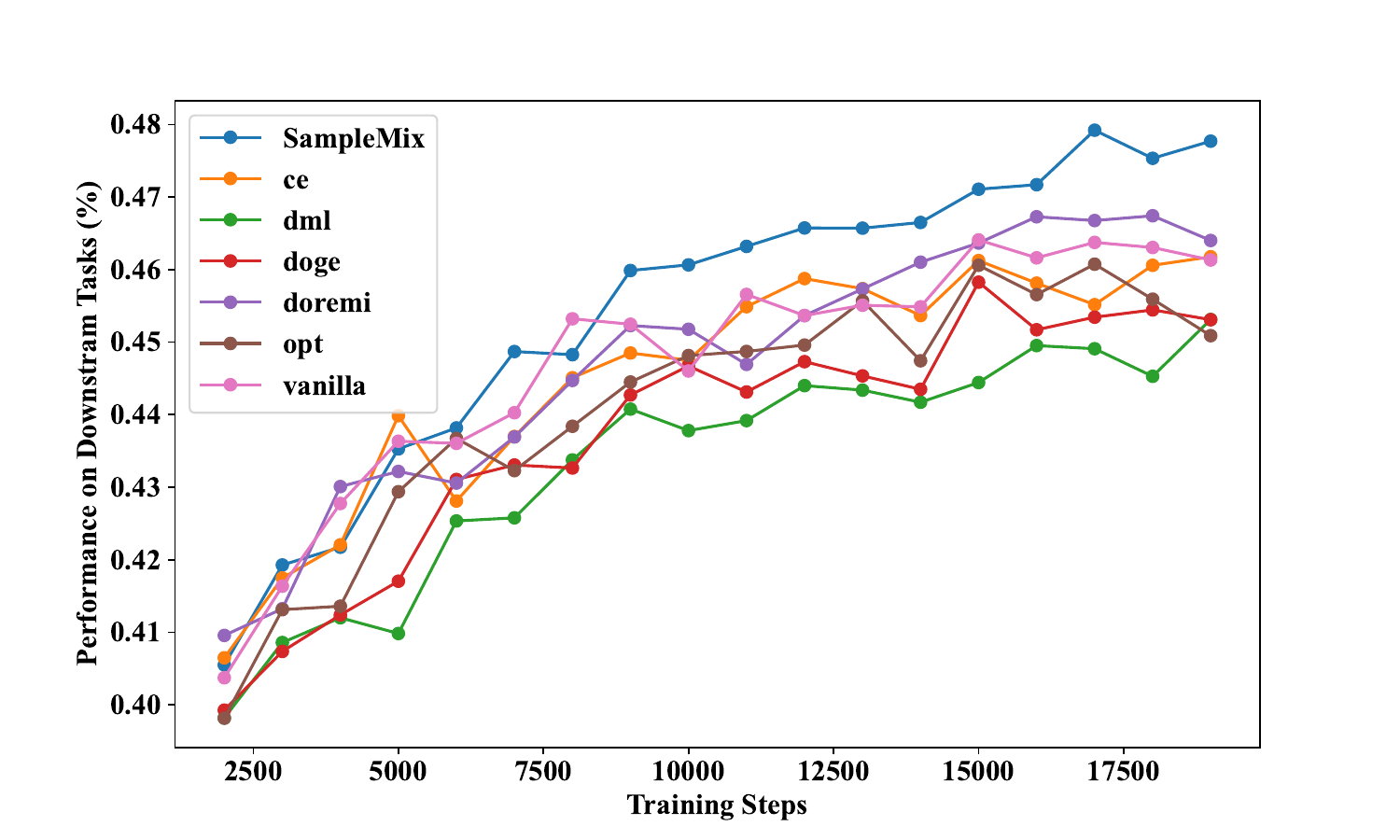}
    \caption{Coverage speed of all baselines and SampleMix. SampleMix achieves the best training efficiency.}
    \label{fig:speed_up_all}
\end{figure*}

Figure \ref{fig:speed_up_all} shows the full comparison of SampleMix and all baselines. SampleMix achieves the baselines' accuracy using 1.4x to 2.1x fewer training steps.

\section{Analysis of Sampling Count Distribution}
\begin{figure*}[!hbt]
  \begin{subfigure}[b]{0.48\linewidth}
    \centering
    \includegraphics[width=\linewidth]{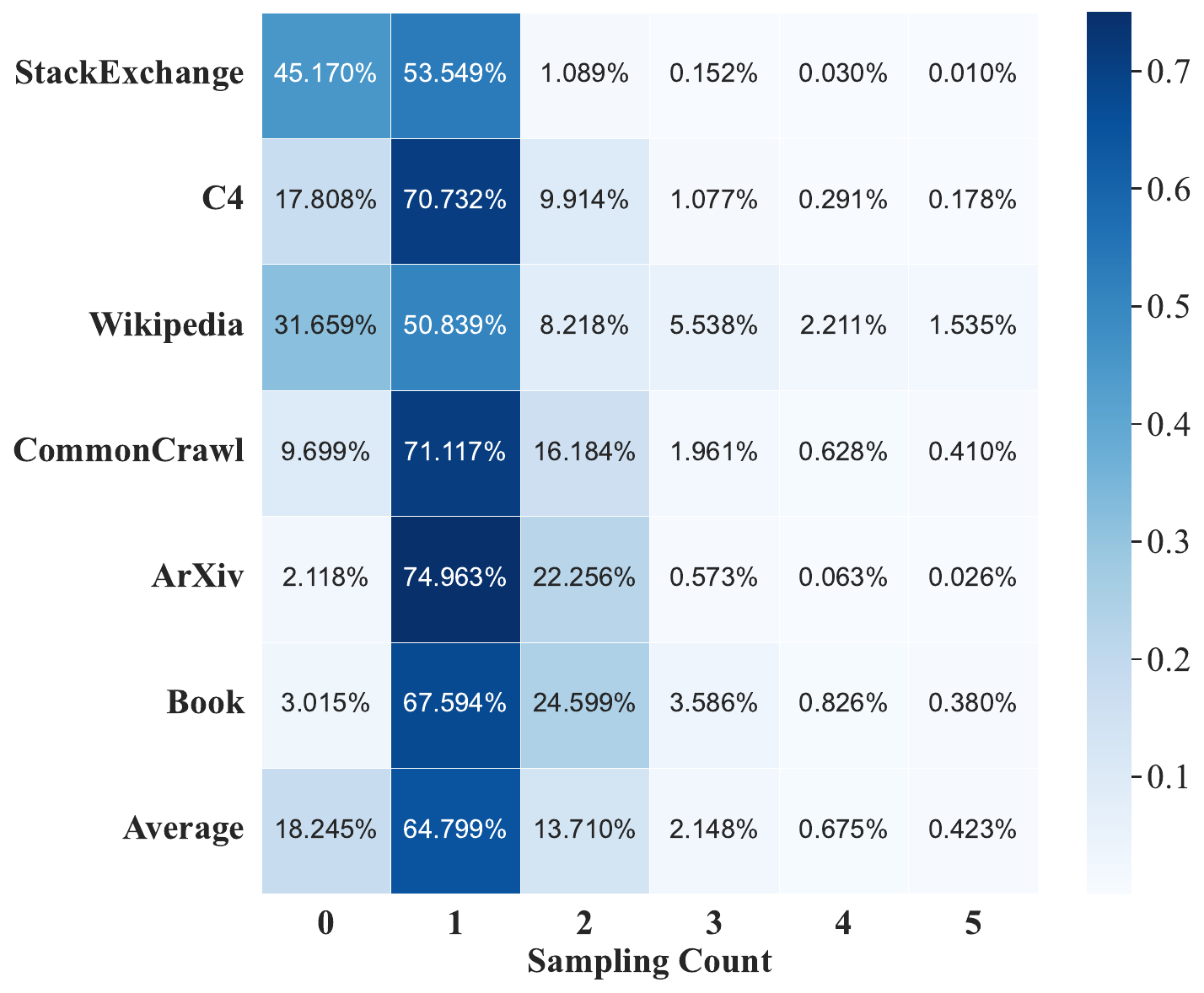}
    \caption{Proportion of different sampling count for $T_{\mathrm{tgt}}=T_{\mathrm{src}}$}
    \label{fig:sampling_count_100B}
  \end{subfigure}
  \hfill  
  \begin{subfigure}[b]{0.48\linewidth}
    \centering
    \includegraphics[width=\linewidth]{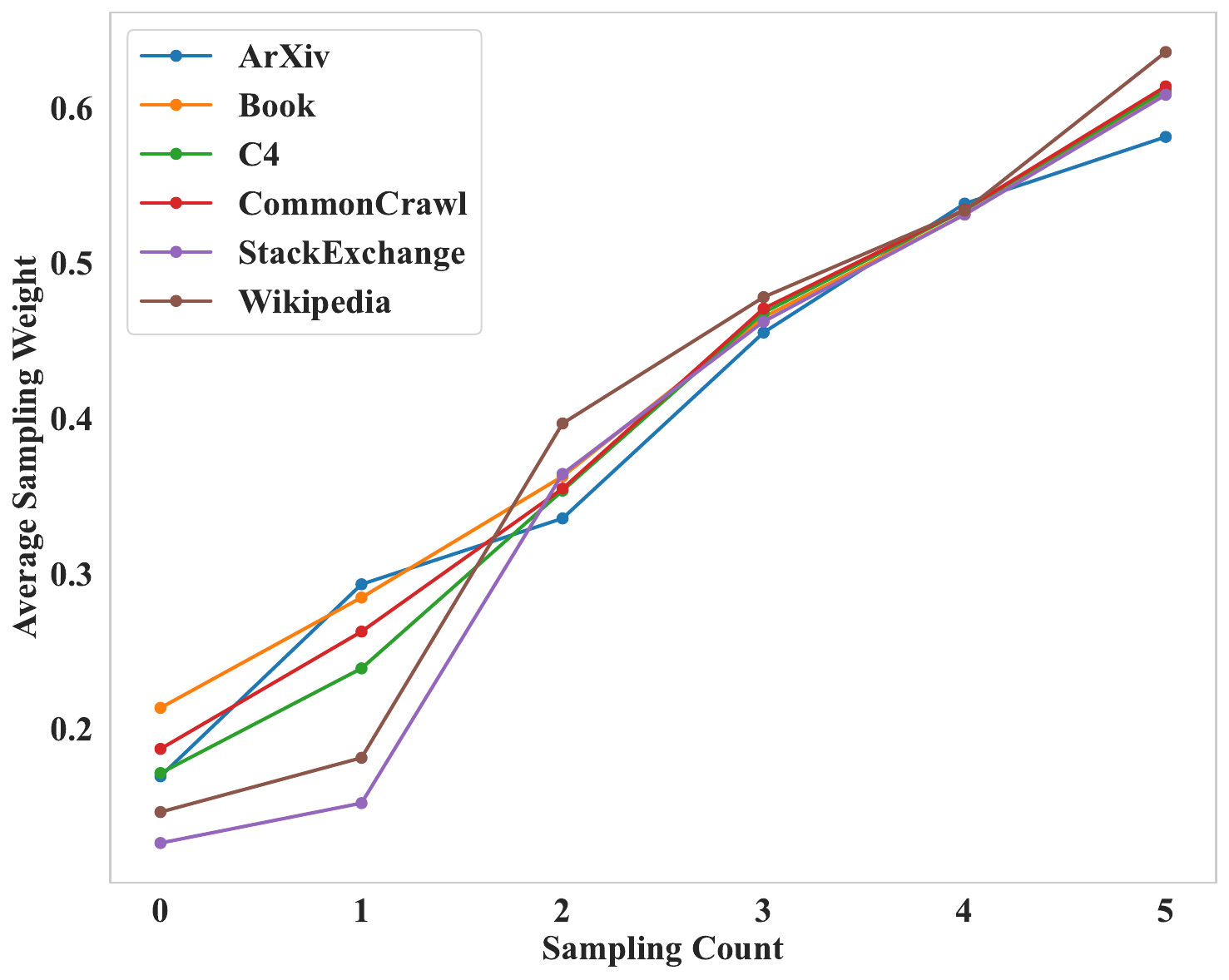}
    \caption{Sampling weight (i.e., $p(x)$) of different sampling counts for $T_{\mathrm{tgt}}=T_{\mathrm{src}}$}
    \label{fig:sampling_score_100B}
  \end{subfigure}
  \caption{Analysis of sampling counts.}
  \label{fig:sampling_count_distribution}
\end{figure*}

Figure \ref{fig:sampling_count_100B} presents the distribution of sampling counts for each domain. Although our target training budget $T_{\mathrm{tgt}}$ is approximately equal to the size of the candidate pool $T_{\mathrm{src}}$, our method strategically discards documents with the lowest quality and diversity by assigning them a sampling count of zero. This approach contrasts with traditional methods that utilize uniform sampling across all documents.
In Figure \ref{fig:sampling_score_100B}, we display the sampling weights corresponding to the sampling counts. The results demonstrate that our method allocates higher sampling counts to samplers with larger sampling weights, aligning with our expectations.
Additionally, the distribution of sampling counts exhibits significant variation across different domains. This variability underscores our method's effectiveness in capturing both fine-grained variations and commonalities among diverse domains, ensuring a more nuanced and efficient sampling process.

\end{document}